\documentclass[letterpaper,twocolumn,10pt]{article}
\usepackage{usenix2019_v3}
\usepackage{amsmath}                                
\usepackage{amssymb}                                
\usepackage{booktabs}                               
\usepackage{mhchem}                                 
\usepackage{graphicx}                               
\usepackage{pgfplotstable}                          
\usepackage[load=addn]{siunitx}                     
\usepackage{subcaption}                             
\usepackage{wrapfig}                                
\usepackage{tcolorbox}                              
\usepackage{xfp}                                    
\usepackage{soul}
\usepackage{makecell}

\DeclareMathOperator*{\argmin}{arg\,min\,}          
\newcommand{\ie}{i.e.,}                             
\newcommand{\eg}{e.g.,}                             
\newcommand{\shortsection}[1]{
    \vspace{1mm}\noindent\textbf{#1.}}
\newenvironment{sidebar}                            
    {\figure[!t]
    \tcolorbox}
    {\endtcolorbox
    \endfigure}

\newcommand{\co}{\ce{^{60}Co}}                      
\newcommand{\na}{\ce{NaI{(Tl)}}}                    
\newcommand{\swabv}{\texttt{MCNP}}                  
\newcommand{\software}{%
    \texttt{Monte Carlo N-Particle Transport Code}} 

\pgfplotstableread[col sep=comma]{
    metrics.csv}{\metrics}

\pgfplotstablegetelem{0}{angles}\of{\metrics}       
\pgfmathsetmacro{\mle}{\pgfplotsretval}                  
\pgfplotstablegetelem{1}{angles}\of{\metrics}       
\pgfmathsetmacro{\rte}{\pgfplotsretval}                  
\pgfplotstablegetelem{2}{angles}\of{\metrics}       
\pgfmathsetmacro{\mlre}{\pgfplotsretval}                  
\newcommand{\error}{37}                          
\newcommand{\rtesd}{, 95\% \textrm{ CI} \pm 0.07}
\newcommand{\mlresd}{, 95\% \textrm{ CI} \pm 0.11}
\newcommand{\mlesd}{, 95\% \textrm{ CI} \pm 0.04}

\pgfplotstablegetelem{0}{distances}\of{\metrics}    
\pgfmathsetmacro{\mlde}{\pgfplotsretval}                  
\pgfplotstablegetelem{1}{distances}\of{\metrics}    
\pgfmathsetmacro{\mldre}{\pgfplotsretval}                  
\newcommand{\mldresd}{, 95\% \textrm{ CI} \pm 0.55}
\newcommand{\mldesd}{, 95\% \textrm{ CI} \pm 0.54}

\pgfplotstablegetelem{3}{angles}\of{\metrics}       
\pgfmathsetmacro{\lmle}{\pgfplotsretval}                  
\pgfplotstablegetelem{4}{angles}\of{\metrics}       
\pgfmathsetmacro{\lrte}{\pgfplotsretval}                  
\pgfplotstablegetelem{5}{angles}\of{\metrics}       
\pgfmathsetmacro{\lmlre}{\pgfplotsretval}                  
\newcommand{\lerror}{26}                              
\newcommand{\lrtesd}{, 95\% \textrm{ CI} \pm 0.22}
\newcommand{\lmlresd}{, 95\%\textrm{ CI} \pm 0.37}
\newcommand{\lmlesd}{, 95\%\textrm{ CI} \pm 0.17}

\pgfplotstablegetelem{2}{distances}\of{\metrics}    
\pgfmathsetmacro{\lmlde}{\pgfplotsretval}                  
\pgfplotstablegetelem{3}{distances}\of{\metrics}    
\pgfmathsetmacro{\lmldre}{\pgfplotsretval}
\newcommand{\lmldresd}{, 95\%\textrm{ CI} \pm 3.66}
\newcommand{\lmldesd}{, 95\%\textrm{ CI} \pm 3.74}

\newcommand{\itembase}[1]{\setlength{\itemsep}{#1}}

\begin{document}

\date{}

\newcommand{\faketitle}[1]{
\twocolumn[
  \begin{@twocolumnfalse}
\vspace{5em}
\centerline{{\bf\large #1}}
\vspace{5em}
  \end{@twocolumnfalse}
  ]
}

\title{\Large \bf
Improving Radioactive Material Localization by \\ Leveraging Cyber-Security Model Optimizations
}

\author{
{\rm Ryan Sheatsley}\\
\and
{\rm Matthew Durbin}\\
\and
{\rm Azaree Lintereur }\\
\and
{\rm Patrick McDaniel}\\
} 

\maketitle


\begin{abstract}

    One of the principal uses of physical-space sensors in public safety
    applications is the detection of unsafe conditions (e.g., release of
    poisonous gases, weapons in airports, tainted food). However, current
    detection methods in these applications are often costly, slow to use, and
    can be inaccurate in complex, changing, or new environments. In this paper,
    we explore how machine learning methods used successfully in cyber domains,
    such as malware detection, can be leveraged to substantially enhance
    physical space detection. We focus on one important exemplar
    application--the detection and localization of radioactive materials. We
    show that the ML-based approaches can significantly exceed traditional
    table-based approaches in predicting angular direction.  Moreover, the developed
    models can be expanded to include approximations of the distance to
    radioactive material (a critical dimension that reference tables used in
    practice do not capture). With four and eight detector arrays, we collect
    counts of gamma-rays as features for a suite of machine learning models to
    localize radioactive material. We explore seven unique scenarios via
    simulation frameworks frequently used for radiation detection and with
    physical experiments using radioactive material in laboratory environments.
    We observe that our approach can outperform the standard table-based
    method, reducing the angular error by \SI{\error}{\percent} and
    reliably predicting distance within \SI{\mlde}{\percent}. In this way, we show
    that advances in cyber-detection provide substantial opportunities for
    enhancing detection in public safety applications and beyond.
    
\end{abstract}

\section{Introduction} \label{introduction}

The integration of computation and sensing has revolutionized the management of
physical spaces~\cite{humayed_cyber-physical_2017}. For example, new
capabilities enable smart buildings that reduce energy use and lessen carbon
footprints, smart homes which ease our personal lives, and smart
infrastructures which support  semi-autonomously secured spaces.  Collectively,
these Cyber-Physical Systems (CPS) are driving massive innovation, and in
particular, advancing public safety in many domains.  Security---the protection
of physical spaces from adversaries who wish to manipulate or harm the space or
those who reside in it---is one of the important areas being advanced.
Specifically, detection of adversarial entities, actions, or dangerous states
is  the principal use of physical-space sensors. 

One of the most well known detection problems in physical security is the
localization of radioactive materials. The use of nuclear technology has grown
since the discovery of radiation~\cite{noauthor_nuclear_nodate} and it is now
found in many applications, including power, medicine, and space.  As the use
of nuclear technology increases, so does the potential for misuse of
radioactive materials. The canonical domain for discussing the detection of
rogue radioactive materials is the container shipping industry (specifically,
cargo inspection). Shipping containers are critical infrastructure, yet
represent an ideal mode of transportation for adversaries due to their low
cost~\cite{congress_of_the_united_states_congressional_budget_office_scanning_2016}
and the presence of large amounts of metal (and other shielding materials)
which attenuates radioactive signals, substantially reducing the efficacy of
radiation detectors to extract relevant signals from the surrounding
noise~\cite{grypp_design_2014}. As yet another example, urban search
applications during large public gatherings (\eg{} the Super Bowl, or Times
Square during New Year's Eve) face similar challenges; the surrounding
buildings can cause severe signal attenuation and impede search effectiveness.

Over the past few decades, a conventional technique for source localization (known as
\textit{Directional Gamma-ray Detection}) has relied on the use of
pre-populated datasets (\ie{} \textit{reference tables}) calibrated at specific
distances in laboratory environments~\cite{schrage_low-power_2013,
hanna_directional_2015}. In a process thematically similar to collecting
malware signatures from known samples, the table is built using location
templates. When a location needs to be screened, gamma ray detectors, placed in
a fixed geometry, are used to acquire counts. The distribution of counts across
the detectors are compared against these reference tables to predict if a
source is present, and the corresponding detector-to-source angle. Even with
these methods, a secondary phase is often necessary, where responders (paired
with portable detectors) manually search for the radioactive material on
foot~\cite{remick_u.s._2005, klann_current_2009, ziock_lost_2002}. A central
limitation to this process is comparing the readings from the detectors to the
reference tables: in non-trivial cases, attenuation, scattering, shielding, and
other naturally occurring phenomena can significantly deviate the
characteristics of the acquired gamma-ray signals.  These factors limit the
utility of reference tables, and are further exacerbated as the true distance
to the radioactive source diverges from the calibration distance of the
reference table.

Localization of radioactive materials has a striking similarity to one of the
most fundamental problems in cyber-security: detection. Similar to malware,
spam, or network intrusion detection, an adversary hides a malicious artifact,
and the challenge falls upon the defender to use information from the
environment (often overburdened by noise) to detect and locate that artifact.
For the past four decades, the security community has developed techniques that
reveal information-rich artifacts and methods to amplify desirable signals from
the surrounding noise. Our insight is that the application of techniques from
cyber-security to radioactive material localization has the potential to
produce significant advancements in physical security.

From a computer security perspective, the physical phenomena described earlier
(\eg{} attenuation, scattering, and shielding) inject noise into the readings.
In fact, the presence of this noise causes naive application of machine
learning to yield results worse than even the relatively inaccurate table-based
approaches. We posit that applying and adapting feature scaling techniques used
in cyber detection domains will mitigate the impact of noise and amplify the 
signal of interest. Specifically, we apply unit norm scaling (regularly used in
spam detection) \& robust feature standardization (often used in domains that
have features with broad scales, \ie{} network intrusion detection) to be able
to discern the signal of interest from the noisy environment.  With these
techniques, we significantly improve the ability to localize radioactive
materials. Further, we explore the abilities of machine learning models to
estimate the distance to radioactive material, thus quantifying both angle and
distance.

One of the new opportunities afforded by the novel application of
cyber-security techniques to this physical domain is that we are no longer
bound to predicting directionality exclusively; we explore the abilities of
machine learning models to estimate the distance to radioactive materials and
show that they are effective in a suite of different environments. The
application and adaptation of these techniques present a  new
capability that has not been achieved for radiation detection applications.
This paper represents a significant step forward in Directional Gamma-ray Detection
with the development of a novel framework to predict distance as well as direction 
(and thus location) with a stationary detection system. 

\begin{figure*}[t]
    \centering
    \includegraphics[width=0.8\textwidth]{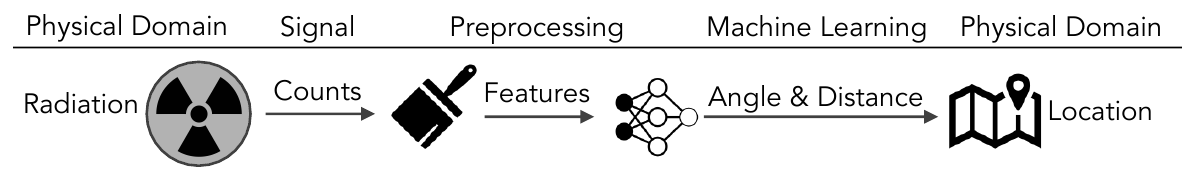}

    \caption{Localizing Radioactive Materials - We apply data curation
    techniques, shown to be successful in cyber-detection applications, to
    counts collected from radioactive signals to predict both the angle to and
    distance from a radioactive source.}

    \label{fig:pipeline}
\end{figure*}

In this paper, we present techniques which have the potential to advance the
current capabilities for locating radioactive materials. We apply and adapt
data curation techniques used successfully in cyber detection domains, and tune
machine learning models to localize radioactive sources. We assess the approach
using the Monte-Carlo-based radiation transport framework
(\software{}~\cite{goorley_initial_2013}) and physical experiments using
radioactive sources in laboratory settings. We design the experiments to
include obstructions that affect radioactive signals, which can serve as a
proof-of-concept for cargo inspection and urban search scenarios. An overview
of our approach is shown in Figure~\ref{fig:pipeline}.

We evaluate this approach with six different models on seven datasets, of which
five are simulated and two are experimental (collected in a laboratory
environment). With the simulated experiments we find that the cyber-inspired
approach reduces the angular error by \SI{\error}{\percent} (\ang{\rte}\(
\rtesd\) from the reference table to \ang{\mle}\(\mlesd\) with our
approach) and we can predict distance within \SI{\mlde}{\percent}\(
\mldesd\) of the source's location (up to \SI{15}{\meter}). For the laboratory
experiments, we reduce the angular error by \SI{\lerror}{\percent}
(\ang{\lrte}\(\lrtesd\) from the reference table to \ang{\lmle}\(
\lmlesd\) with unit norm scaling) and predict distance within
\SI{\lmlde}{\percent}\(\lmldesd\) of the radioactive materials (up to
\SI{3}{\metre}). Our contributions are:

\begin{itemize}
\itembase{0pt}

    \item We present techniques adapted from cyber-security detection to
        exploit the use of gamma-ray signals for accurate radioactive material
        localization.

    \item We demonstrate that our approach surpasses the traditional
        table-based approach, incurring an average angular error of
        \ang{\mle}\(\mlesd\) vs. \ang{\rte}\(\rtesd\), respectively.

    \item We extend the standard definition of localization to include
        distance. Here, the posited approach can predict distance within
        \SI{\mlde}{\percent}\(\mldesd\) of a simulated radioactive source
        when the source strength is known.

    \item We perform experiments with real, radioactive sources to validate our
        findings in complex laboratory environments. Our approaches surpass the
        table-based method, incurring an average angular error of
        \ang{\lmle}\(\lmlesd\) vs. \ang{\lrte}\(\lrtesd\),
        respectively.  Moreover, we can predict distance within
        \SI{\lmlde}{\percent}\(\lmldesd\) of real radioactive materials.

    \item We provide seven new datasets (including simulated and real data)
        that we make public for future research in this important domain,
        curated for use with machine learning.

\end{itemize}

\section{Problem Definition} \label{problem}

\shortsection{Threat Model} Our problem is essentially a game of hide and seek:
an adversary places a radioactive source and the objective, as the defender, is
to confirm its existence and determine its location. We assume a complex
environment, that is, it contains obstructions (\eg{} buildings) that interfere
with (and thus, obfuscate) the signal produced by the radioactive source.
Further, we assume a stationary environment: the adversary is non-adaptive and
the radioactive source and the radiation detector are stationary (however, the
distance and angle between the source and the radiation detector can change
across different experimental scenarios), as are the surrounding obstructions
in the environment. We also assume that the adversary has no ability to
intervene with the operation of the radiation detector and that it is operating
optimally (and therefore trust the produced readings to be as accurate as
physical phenomena permit).

To some extent, we also assume no a priori knowledge of the radioactive
material being used by the adversary (which we detail in
Section~\ref{approach}). While the experiments are done exclusively with one
material (due to availability, safety, and applicability), the method with
which we detect and localize a radioactive source is agnostic to any gamma ray
emitting isotope used by an adversary. The intuition behind this is
straightforward: while different gamma ray emitting isotopes exhibit unique
radioactive signatures, they all emit quanta within certain energy regions.
Therefore, for this approach, detecting a particular isotope is simply a
function of which portion of the energy region is scanned. Thus, a takeaway of
this work is that responders can be ``blind'', in some sense, to the specific
isotope used by an adversary. 

\shortsection{Detector Setup} The detection of a radioactive source relies upon
the use of a detector array. The radiation
produced by the source interacts with the detectors to generate a signal. This
signal is then captured and used to produce a histogram corresponding to the
energy deposited via different interactions.  The detection system is comprised
of four or eight detectors, each of which collect individual energy histograms.
Here we sub-sample counts over a range of energies which are the inputs (\ie{}
features) to the machine learning models. 

\begin{figure}[t]
    \centering
    \includegraphics[width=0.7\columnwidth]{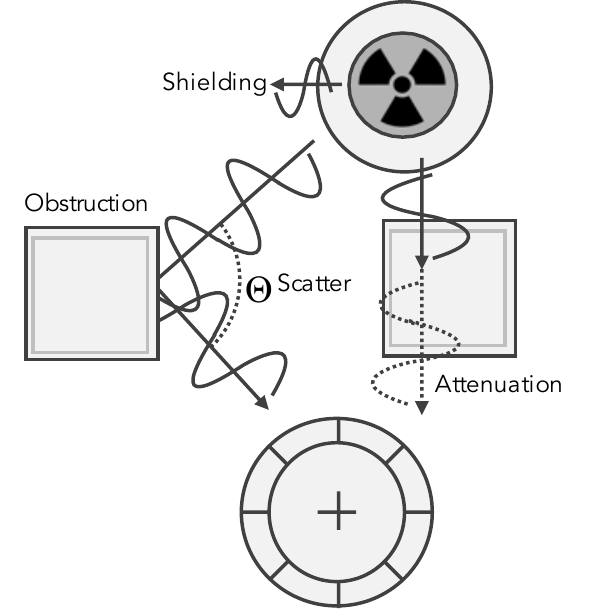}

    \caption{Detection in a complex environment - The
    principal objective is to detect and localize radioactive sources in
    complex environments that induce undesirable phenomena, \eg{} shielding,
    attenuation, scatter, and background noise.}

    \label{fig:environment}
\end{figure}

As mentioned in Section~\ref{introduction}, there are a suite of environmental
factors that can negatively affect the readings of the detector (aiding the
adversary). Figure~\ref{fig:environment} highlights some of the main sources of
these environmental factors, and how they impact the signal. Notably, there are
three central phenomena produced by obstructions: attenuation, scatter, and
shielding. We describe in more detail how these phenomena affect the readings
in Section~\ref{radiation}, but for the purposes of the problem definition,
these broadly just reduce the signal or amplify noise.

\shortsection{Machine Learning for Security} Machine learning has been
successful in computer security detection applications including network
intrusion detection, malware, and zero-day
vulnerabilities~\cite{kubat_machine_nodate, sommer_outside_2010,
tsai_intrusion_2009}. However, machine learning has not been applied to
radioactive source search scenarios, which shares many parallels with domains
within computer security. A central factor for deploying machine learning in
security-sensitive domains is feature scaling~\cite{protic_anomaly-based_2018,
tahir_efficacy_2019, kotagiri_robust_2007}. In network intrusion detection,
there are many different kinds of features, which can contain outliers that can
negatively affect standardization~\cite{kotagiri_robust_2007}. We find that
detecting radioactive materials faces a similar burden, in that the
distributions of gamma-ray counts can contain strong outliers (\ie{} sources of
noise). By applying robust feature standardization techniques that account for
these outliers, we improve the accuracy of many learning algorithms.

As a second optimization, we take inspiration from techniques used
traditionally for spam detection: unit norm scaling. Specifically, we observe
that much like analyzing the relative frequency of words in emails, learning
algorithms are likely to be more accurate in localizing radioactive materials
with relative detector counts rather than gross signals.

\section{Radioactivity in the Physical World} \label{radiation}

Radiation, which is energy in transit, can be either electrically charged
(\eg{} electrons, protons, and alpha particles) or uncharged (gamma rays,
x-rays, and neutrons). Uncharged radiation poses unique detection challenges,
but is not easily shielded~\cite{knoll_radiation_2000}. In this work,
gamma-rays are the principal phenomena of interest as they are not readily
shielded by thin metals (\ie{} shipping containers), unlike charged
particles~\cite{u.s.nrc_nrc:_nodate}. Also, gamma-rays produce unique energy
signatures~\cite{johns_physics_1983} which can be used to classify the
radioactive source, analogous to signatures produced by malware in intrusion
detection systems. Generally, gamma-rays that pass through a detector interact
in one of three ways: the photoelectric effect, Compton scattering, or pair
production. These interactions produce readings that eventually become the
features of the approach.

\shortsection{Poisson Statistics} Absent of physical phenomena (and any
detector deficiencies), one of the most fundamental challenges in interpreting
detector readings is that they are burdened by Poisson statistics. The
underlying uncertainty complicates accurate interpretation of the readings; the
stochastic nature of radioactive decay means that \textbf{the exact same
experiment repeated twice in a row will yield different results}. This fact
gives a fundamental insight into what makes localizing radioactive sources a
challenging problem.

In the simplest scenario (no obstructions, line-of-sight to the radioactive
source, and ideal detector characteristics), the phenomena above, coupled with
Poisson statistics, can have a notable affect on detector results, which
table-based analysis approaches have difficulty rectifying.  Much like network
adversaries who obfuscate their signature to frustrate detection systems by
mixing benign requests in the midst of malicious ones, noise produced by
scattering and attenuation (as well as the fundamental uncertainty) can have a
non-trivial negative effect on this approach with respect to localizing
radioactive sources.

\subsection{Existing Approaches}

The two phase search procedure, which relies on detection and localization, is
a demanding process, both in time and labor, given that responders must
triangulate the radioactive source manually. Currently deployed techniques aim
to combine both phases: by analyzing minute differences between counts received
across detectors in an array of fixed geometry, the angle to the radioactive
source can be determined (within some error). This problem is known as
\textit{directional gamma-ray detection}.  Most conventional techniques use
pre-populated datasets (\ie{} \textit{reference tables}) of known source
locations, calibrated at a specific distance in lab
environments~\cite{schrage_low-power_2013} (shown in
Figure~\ref{fig:reference_table}). However, these approaches suffer in
non-trivial cases where attenuation, scattering, shielding, and other naturally
occurring phenomena affect the detected signals. More, these methods still
require responders to manually search in the suspected direction of the
radioactive source. Thus, they are susceptible to human-error, inaccuracies of
the reference tables, and are bound by the number of responders that can be
equipped with portable detectors to triangulate the source.

For this work, we measure the net counts for each individual detector
and compute their differences to predict both angle and distance (which gives
us localization). This combines two previously disjoint stages, enabling
responders to quickly identify the location of the radioactive material. We
hypothesize there is latent information that characterizes the environment
which enables the location of the radioactive material to be determined.
Consider that if one detector receives more counts than another, it is likely
the source is in the direction of the detector with the highest counts.
However, if an obstruction is directly in front of this detector, then the
neighboring detectors may receive more counts.  The challenge here is to
capture these subtle situations--tools used in detection in computer security
environments are effective at pulling out this embedded information, and we
exploit this observation in our analysis.

\section{Approach} \label{approach}

Localizing radioactive materials in noisy environments shares many of the same
challenges observed in the cyber-security detection space. Here, we briefly
detail the radioactive source used, some relevant characteristics of the 
detector, the framework used in the simulated experiments, describe the feature
scaling adaptations, and present the machine learning algorithms used. 

\subsection{Material Detected}

Cobalt-60 (\co{}) is the radioactive isotope used in the simulations and
laboratory experiments. Cobalt-60 is a relevant isotope to study, as it can be
found in many domains, including medicine, industry, food, and nuclear
power~\cite{international_atomic_energy_agency_gamma_2006, walker_epa_nodate,
malkoske_cobalt-60_nodate}. The widespread use of \co{} means that, in
practice, it is an isotope responders often wish to locate.. Most importantly, while we use \co{} in the 
experiments, we emphasize that these techniques are not specific to this
isotope; many radioactive isotopes have characteristic peaks similar to \co{},
simply at different energies~\cite{knoll_radiation_2000}.

\subsection{The Detectors}

In gamma-ray spectroscopy, there are two main detector types: scintillators and
semiconductors. Though most semiconductors offer better resolution and improved
intrinsic efficiency (\ie{} a high probability of interaction with gamma-rays),
they are expensive, and some requiring cooling to liquid nitrogen temperatures
(\SI{-196}{\celsius})~\cite{saha_physics_2001}. Such requirements were
impractical for this work.

Thus, we use thallium-doped sodium iodide (\na{}) scintillation detectors,
popular in many field applications~\cite{mukhopadhyay_networked_nodate}. There
are a handful of properties that make \na{} detectors useful for
experimentation, namely: room-temperature operation, high efficiency, and large
photofraction (\ie{} the fraction of incident photons fully absorbed). The
popularity and accessibility of the detector makes it an attractive choice for
evaluating the applicability of this approach.

\subsection{Monte Carlo N-Particle Transport Code}

\begin{figure}[t]
    \includegraphics[width=\columnwidth]{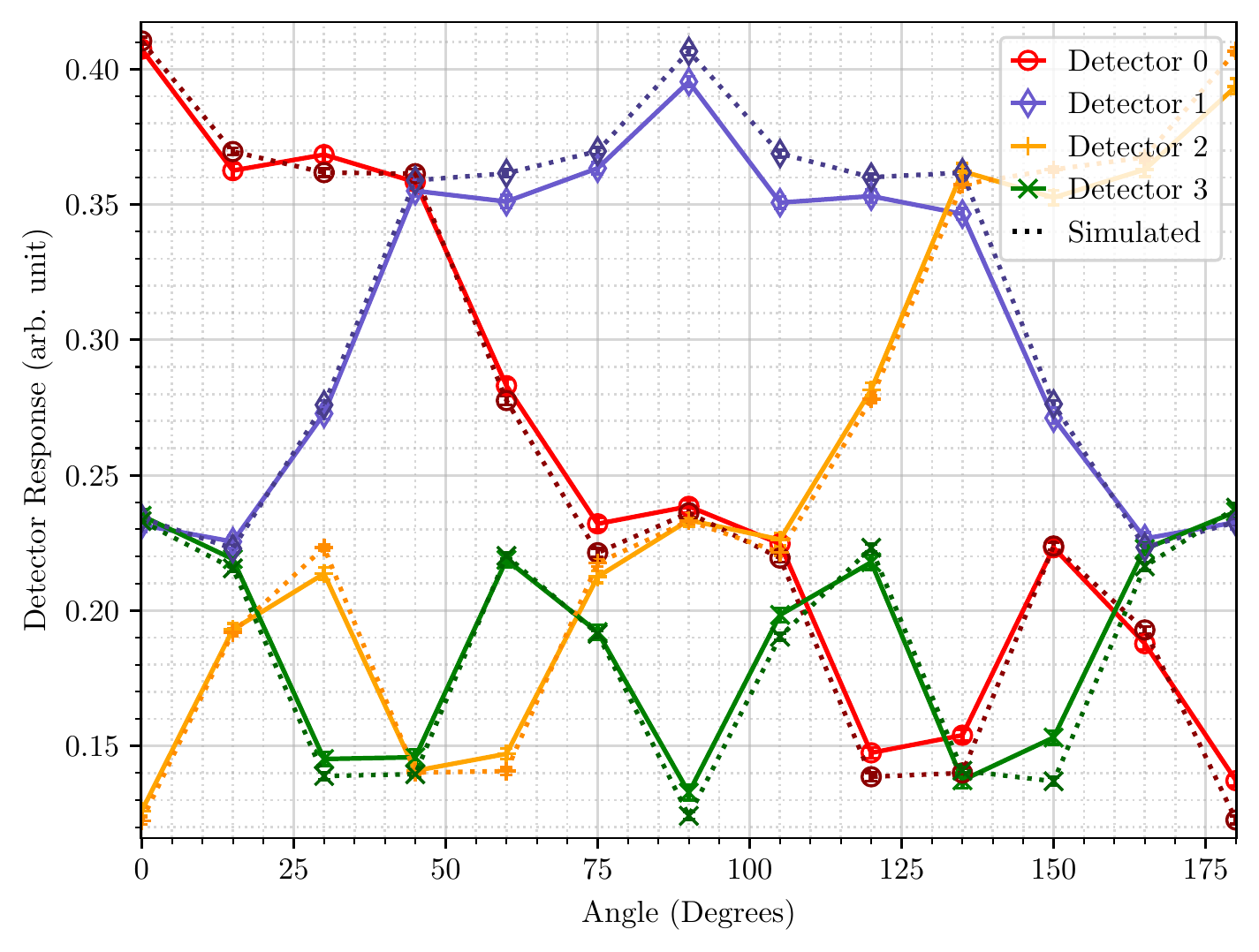}

    \caption{Detector responses as a function of angle for laboratory
    experiments and simulations - \swabv{} is capable of modeling radioactive
    phenomena with striking accuracy and precision. Error bars on response values are included, but comparable in size to the plot markers. \swabv{} is the de facto tool
    used for radiation analysis, particularly large-scale experiments that are
    difficult to safely execute in live laboratory settings.}

    \label{fig:simulation_v_laboratory_measurements}
\end{figure}

\textsc{Monte Carlo N-Particle Transport Code} (\swabv{}) is a Monte Carlo method simulator for radiation
transport\footnotemark{}. It uses Monte Carlo methods to simulate interactions
(\ie{} absorption, and scattering) as radiation propagates through a medium.
Monte Carlo methods are considered to be the de facto standard for applications
in radiation analysis due to their ability to accurately model radiation
transport and interactions~\cite{goorley_initial_2013}. Due to its ability to simulate
nearly any environment, \swabv{} is used in
many fields, including medicine, detector design, reactor design, radiography,
material penetration tests, radiation dosimetry, among
others~\cite{goorley_initial_2013}.
Figure~\ref{fig:simulation_v_laboratory_measurements} showcases the capacity
of \swabv{} to model real radioactive phenomena--the simulated detector
responses map nearly 1:1 onto the laboratory detector readings. Differences 
between simulated and laboratory readings are largely attributed to the effects 
of \textit{gain-shift}~\cite{casanovas_temperature_2012, reeder_performance_nodate}, 
stemming from temperature and other environmental influences. Gain shift has the effect of ``shifting'' the 
spectra, which can lead to counts artificially being added or subtracted to the reading. 
As each detector experiences differing amounts of gain shift, these small effects can 
propagate to notable differences in the normalized input features. To partially mitigate this, energy calibrations are regularly performed.

\footnotetext{``Radiation Transport'' software simulates the propagation of
radiation through space and its interactions in media.}

\subsection{Feature Selection}

With a popular isotope and effective detector, we return to the central goal:
localization of radioactive sources in complex environments. To achieve this 
goal, we first ask a basic question: \textit{what can we measure?} We defer to
historically successful techniques to answer this question.

As described in Section~\ref{radiation}, prior work has combined detection and
direction (\ie{} angle) into a single phase by analyzing slight variances in
the responses of each detector in a fixed geometry
array~\cite{durbin_development_2019}. Here, we use gamma-ray counts as
features, and extend the analysis to localizing (\ie{} predict both angle and
distance) radioactive sources. In this way, we seek to combine all phases of detection
and localization into one step, averting cost for any special equipment and
saving time by avoiding a manual foot search.

\subsection{Feature Optimization \& Cyber-security}

The abundance of research in localizing radioactive sources has many answers to
the question asked above. However, there is yet another question that we must
ask: \textit{how do we filter noise without dampening the signal of interest?}
To answer this question, our intuition leads us to our central hypothesis:
techniques used in the cyber-security detection space can be useful in
localizing radioactive materials. We describe the techniques and relevant
adaptations below\footnotemark{}.

\footnotetext{It is worth noting that we tried other data manipulation
techniques that had marginal (or sometimes even negative) effect on the
accuracy of the models, namely: 0-1 rescaling, normalization (\ie{} mean
centered at 0), and standardization (\ie{} mean centered at 0 and a standard
deviation of 1).}

\shortsection{Robust Feature Standardization} For many learning algorithms,
standardizing feature scales is an important prerequisite. A common technique
is to subtract the mean and scale to unit variance. However, standard
techniques can be negatively affected by features that have highly skewed
distributions, like those seen in network intrusion
detection~\cite{kotagiri_robust_2007}. Instead, subtracting the median and
scaling features according to some quantile range has been shown to produce
better results for features with outliers. We standardize the features via:

\begin{equation}
    \hat{x} = \frac{x - \tilde{x}}{\max{Q_3} - \max{Q_1}}
\end{equation}

\noindent where \(x\) is the original feature, \(\hat{x}\) is a standardized
feature, \(\tilde{x}\) is the median value, and \(\max{Q_i}\) is the maximum
value for the \(i^{\textnormal{th}}\) quantile. By scaling based on the maximum
value in some quantile, we mitigate the negative influence outliers may have on
the accuracy of some learning techniques.

\shortsection{Unit Norm Scaling} \(l_p\) normalization is an arguably uncommon
feature scaling technique (\(l_2\) \textit{regularization} on model parameters
is a common use of \(l_p\) norms in machine learning). If we represent a
dataset as an \(M\times N\) matrix of \(M\) samples and \(N\) features, then
most feature scaling techniques operate across \textit{all samples} (\ie{}
\(M\times 1\))--that is, one particular feature for all samples is scaled in
some manner. However, unit norm differs in that it operates across \textit{one
sample} (\ie{} \(1\times N\)). We can formulate unit norm scaling as:

\begin{equation}
    \hat{x} = \frac{x}{\|x\|_{l_p}}
\end{equation}

\noindent where \(\hat{x}\) is a \(l_p\) norm scaled input. Our intuition for
using unit norm scaling follows successful applications of natural language
processing towards spam detection: a natural objective for spam detection is to
determine the \textit{term frequency} of words (or n-grams) in an email. For
example, it is difficult to draw any conclusions if the bigram ``free money''
appears a handful of times in email. However, stronger conclusions can be drawn
if ``free money'' was the \textit{most common} bigram. We apply this same
reasoning to predicting the angle to a radioactive source: whether or not a
detector receives \si{10} or \si{100} counts is hardly useful; what is more
important is that a detector received \textit{the most} counts (which is then
the most likely direction to the radioactive source). This insight leads to
substantial increases in angle prediction accuracy.

\subsection{Reference Tables \& Machine Learning}

Here, we describe the reference tables and learning algorithms used in this work.  
The techniques presented here were specifically chosen, as they offer unique
advantages over one another, such as interpretability, scalability, and accuracy.
Further details are presented in Appendix~\ref{appendix}.

\begin{figure}[t]
    \includegraphics[width=\columnwidth]{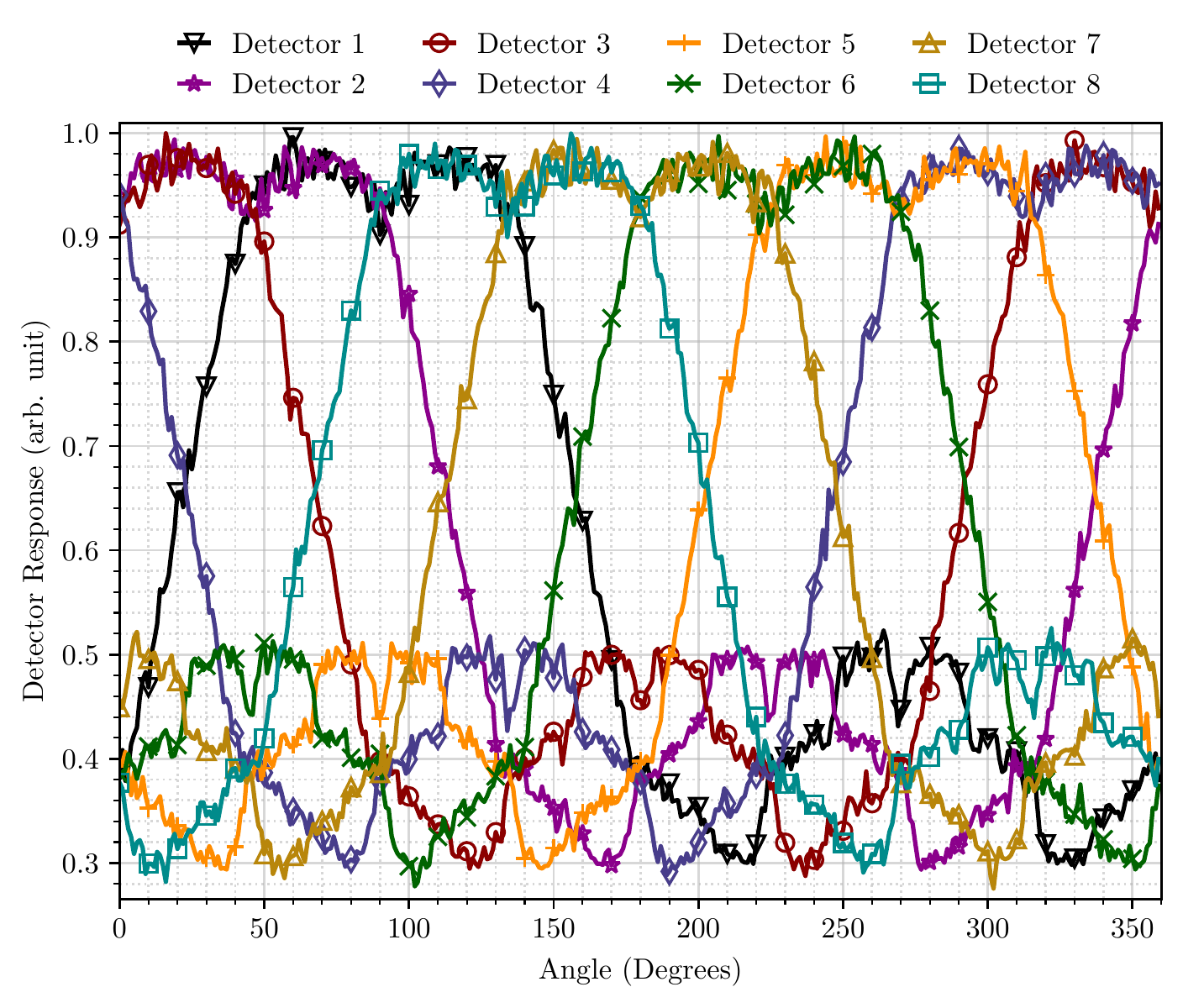}

    \caption{\swabv{} simulated detector responses as a function of source
    angle - reference tables are built from these responses and are then used
    to approximate unknown sources. Calculated error bars are comparable in 
    size to line thickness, and are thus excluded.}

    \label{fig:reference_table}
\end{figure}

\shortsection{Reference Tables} As described in Section~\ref{radiation},
reference tables are commonly used for directional gamma-ray detection.  To
build a reference table: detectors are setup in a fixed geometry, a known
source is selected and placed at a fixed distance from the detector array, and
the relative counts for each detector are recorded at varying source angles.
These reference tables are ostensibly a closed-form approximation of the
phenomena as they encode the response of a detector as a function of angle to
the source, as shown in Figure~\ref{fig:reference_table}.

Once the reference table is calibrated, the responses of the detectors to an
unknown source are compared to the reference table. Often with a least-squares
regression analysis, where the angle to the unknown source is predicted by:

\begin{equation} \label{eq:lsr}
    \theta_{\gamma} = \argmin_{\theta} (\Gamma(\theta)-x)^2
\end{equation}

\noindent where \(\theta_{\gamma}\) is the angle predicted by the reference
table, \(\Gamma(\theta)\) are the calibrated detector responses for some angle
\(\theta\), and \(x\) is a vector of detector counts for a particular
observation. Equation \ref{eq:lsr} leads us to two observations: 1) differences
between the calibration environment and real world environment (such as the
presence of obstructions) will lead to discrepancies in the relative detector
responses, potentially leading responders in the wrong direction, and 2) often
more severely, slight variations in distance can produce profoundly different
distributions of counts than any observation for which the reference table was
calibrated. 

\shortsection{Logistic Regression} Logistic regression (LR), akin to linear
regression, computes a weighted sum of input features with an additional bias
term, and applies the logistic function to this
sum~\cite{wright_logistic_1995}. While more sophisticated techniques have
emerged, each have limitations. We include logistic regression models in this 
work to investigate if simpler models suffice to perform localization tasks
accurately and quickly.

\shortsection{Support-Vector Machines} Prior to the inception of deep learning,
support-vector machines (SVMs) dominated machine learning benchmarks across
many domains~\cite{boser_training_1992}. SVMs are attractive as they can form
non-linear decision boundaries, which may be necessary given the noisiness of
this domain.

\shortsection{k-Nearest Neighbors} k-Nearest Neighbors (kNN) is a
non-parametric approach~\cite{altman_introduction_1992}. As the observed count
distributions in this domain can change rapidly in a variety of unique
environments, kNN is particularly useful as it does not make any assumptions
about the underlying data. 

\shortsection{Decision Trees} Decision Trees (DT) are flexible machine learning
algorithms that are commonly used today~\cite{quinlan_induction_1986}. They
require minimal data preparation, have low performance overheads, and offer
intuitive explanations of the formed decision boundaries. Decision trees are
appealing in this domain not only as another non-parametric technique, but also
because of their white-box design: the learned decisions are easy to interpret,
which can be useful in understanding subtle changes in the decision process as
a function of the environment.

\shortsection{Deep Neural Networks} Deep neural networks represent a
state-of-the-art class of learning techniques that have demonstrated success in
the most challenging machine learning benchmarks~\cite{fine_feedforward_1999}.
Definitions vary, but generally speaking, deep neural networks often refer to
any class of artificial neural networks with multiple layers between the input
and output layers. For this work, the ``deep neural network'' is a
fully-connected feedforward network, with some number of hidden layers, trained
with back-propagation, using the rectifier (colloquially, ``ReLU'') as the
activation function.

\section{Evaluation} \label{evaluation}

In this section, we evaluate the approach on seven datasets: five were
simulated and two were measured\footnotemark{}. The experiments were performed
on a Dell Precision T7600 with Intel Xeon E5-2630 and NVIDIA Geforce TITAN X.
We used Scikit-learn~\cite{pedregosa_scikit-learn_2011} for data curation and
for instantiating the machine learning models.  We ask:

\footnotetext{The radioactive \co{} source used in the laboratory experiments
was a low activity source. At approximately \SI{1}{\micro\curie}, it is safe to
handle with the appropriate safety procedures, which were defined and strictly
followed for all the measurements.}  

\begin{itemize}
\itembase{0pt}

    \item Is the naive machine learning solution sufficient for localizing
        radioactive sources?

    \item Can our optimizations approximate or exceed the performance of
        existing table-based angle predictions?

    \item When source strength is known, can distance be predicted and, if so,
        how is the accuracy affected as a function of distance from the
        radioactive source?

\end{itemize}

\noindent {\bf Summary}: The simulation and
physical laboratory experiments demonstrate that the developed techniques outperform the
reference table for predicting angle by \SI{\error{}}{\percent} (down from
\ang{\rte} to \ang{\mle}) and estimate distance within \SI{\mlde}{\percent} of the
distance to a radioactive source.

\subsection{Experimental Scale \& Parameters}

The simulated and laboratory datasets contain a radioactive source that is located 
\SIrange[range-units=single]{1}{15}{\metre} and
\SIrange[range-units=single]{1}{3}{\metre} away from the detector array center, respectively. Simulated
acquisitions were approximated as \SI{14}{\second} counts of a \si{1} and
\SI{10}{\micro\curie} \co{} source, while laboratory acquisitions were 5 minute
counts of an approximately \SI{1}{\micro\curie} \co{} source. With scaling
(discussed below), these are common parameters in this field, and serve as a
proof-of-concept for this analysis method. 

The experiments contain three factors that are scaled in a manner that makes
the datasets especially challenging: attenuation, measurement time, and
distance. These datasets serve as benchmarks for evaluating approaches on real
radioactive material subject to all forms of physical phenomena and
environmental factors.

\shortsection{Attenuation} Radioactive sources used by medical or industrial
entities can be on the order of \SI{\sim 100}{\curie} (around eight orders of
magnitude stronger than the real source used in this
work)~\cite{international_atomic_energy_agency_gamma_2006,
ortiz_lessons_nodate}. These sources have the potential to be used by the
adversary, and thus have activities similar to those which these methods could be
used to localize. Separate simulations were conducted with the four detector
array and \co{} sources of various activities to gauge the scaling of
obstruction thickness and attenuation. Results showed that the attenuation
effect of a single cinder block (\SI{10} {\centi\meter} of solid concrete) on a
\SI{1}{\micro\curie} source is approximately the same as the attenuation effect
of \SI{150} {\centi\meter} of solid concrete on a \SI{1}{\curie} source. Thus,
the experiments model a challenging scenario where a weak radioactive source is
in a thick concrete building.

\shortsection{Sample Time} We also modeled the measurement time in the
simulated datasets to be comparable to the times used in realistic search
scenarios~\cite{knoll_radiation_2000}. Recall, radioactive sources decay at a
particular \textit{rate}---that is, a \SI{1}{\curie} source sampled for 5
minutes will produce (theoretically) identical results to a \SI{1/2}{\curie}
source sampled for 10 minutes (assuming a half-life much greater than the
measurement time). Since these sampling times are comparable to realistic
scenarios, but with a source potentially 8 orders of magnitude weaker than
those in real scenarios, predicting angle and distance is challenging from a
scaling perspective.

\shortsection{Distance} Most sources radiate in an isotopic manner, and thus
fall subject to many laws in signal processing, notably the inverse-square law.
This means that the intensity perceived by a detector decreases squarely with
distance. Additional simulations were conducted to gauge the scaling effects on
distance, to put the laboratory measurements in perspective to real world
expectations. Results isolating the effects of geometry alone showed that the
counts we receive with the \SI{1}{\micro\curie} \co{} source in five minutes at
\SI{3} {\meter} are comparable to the counts we would get from a \SI{1}{\curie}
\co{} source in a single second at \SI{100} {\meter}. So while the sources used
in the laboratory experiments are challenging to detect at 0.5 to
\SI{3}{\meter} away  (with short data acquisition times), they more than scale
to parameters useful for actual source search applications.

\subsection{Datasets}

The first task was to create datasets of  simulated and real-world measured
gamma-ray counts in various settings to enable evaluation of the detection
algorithm.  We generated seven, which we will refer to as
\textit{S-Dataset 1, S-Dataset 2 \(10^{6/7}\), S-Dataset 3 \(10^{6/7}\)} for
the simulated experiments (from \swabv{}), \textit{L-Dataset 1}, and
\textit{L-Dataset 2} for the laboratory experiments. The \(10^6\) and \(10^7\)
variants of S-Dataset 2 and 3 describes the number of simulated gamma-rays per
trial. The different numbers of simulated gamma-rays correspond to different
source strengths or measurements times, and represent differing levels of
statistics. All scenarios use the same radioactive source (\co{}) at varying
distances, angles, and obstruction locations, as described below and summarized in Table~\ref{table:dataset_details}.

\shortsection{Simulated Datasets} The datasets were generated with an
\si{8}-detector array and a source at varying distances between \si{1} and
\SI{15}{\meter}. S-Datasets \si{1} \& \si{2} contain \si{72,000} samples where
the source is uniformly rotated a full \ang{360} around the detector array (at
roughly \ang{1} increments), while S-Dataset \si{3} contains \si{27,000}
samples and the radioactive source is only rotated \ang{90}. For each trial,
either \(10^6\) or \(10^7\) gamma-rays were simulated, corresponding to a
\SI{14}{\second} count of a \SI{1}{\micro\curie} or a \SI{10}{\micro\curie}
\co{} source. S-Datasets \si{2} \& \si{3} contained a solid concrete
obstruction that mimics the effects of a concrete building. In S-Dataset
\si{2}, the obstruction was stationary; in S-Dataset \si{3}, the obstruction
was randomly placed between ten locations.  
Table~\ref{table:dataset_details} provides experiment details.

While one million gamma-rays may sound significant, recall the isotropic nature of
radiation and the variety of physical phenomena described in
Section~\ref{radiation}; in reality, less than \SI{0.1}{\percent} of these
gamma-rays will cause some interaction (either positively or negatively) with
the detector array \SI{3}{\meter} from the source.

\begin{figure}[t]
    \centering
    \includegraphics[width=0.98\columnwidth]{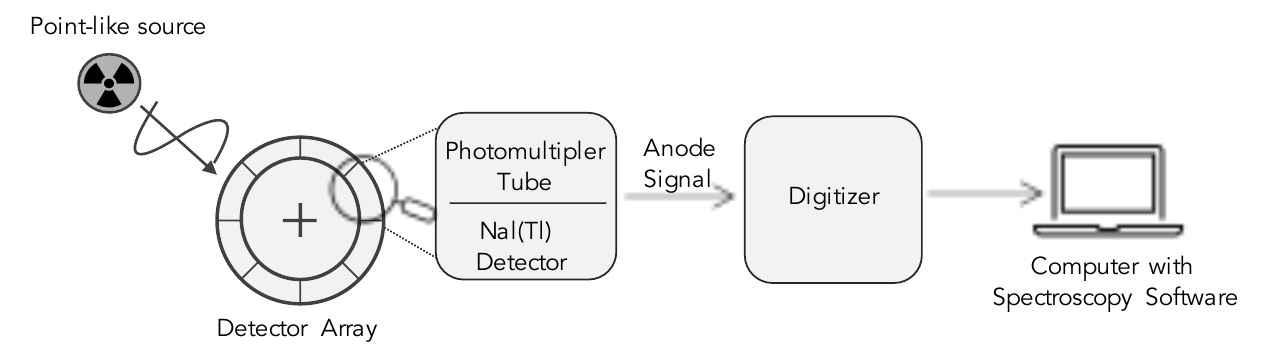}

    \caption{Detection in Radioactive Environments - The gamma-rays from the \co{} source interact
    with the detector array, consisting of \ce{NaI} detectors which emit light upon gamma-ray interactions, photomultiplier tubes (PMTs) to convert the light into an analog pulse, 
    and a digitizer to convert the analog pulse into a digital signal. The digital
    signal is then processed by spectroscopy software to convert the
    signals into the ``counts'', which are used as input to the machine 
    learning models.} 

    \label{fig:block_diagram}
\end{figure}

\begin{figure}[t]
    \includegraphics[width=\columnwidth]{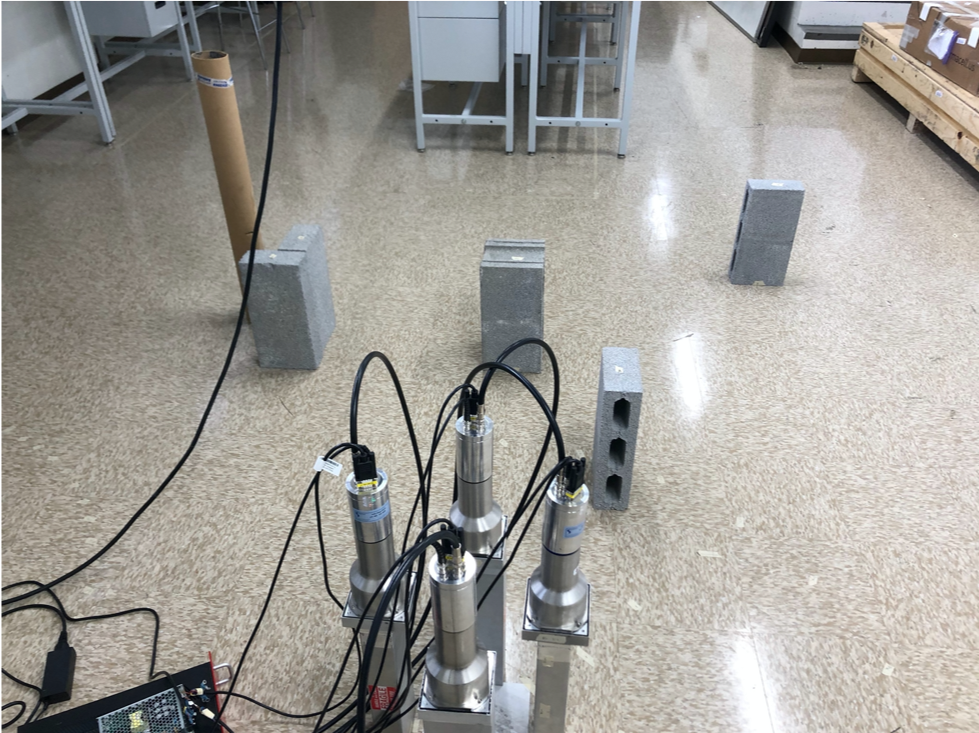}

    \caption{Setup of L-Dataset 2 - This is a photograph of the detector array
    in the laboratory. The \co{} source is attached to the cardboard tube. The
    concrete blocks are the obstructions and cause the behavior of the
    gamma-rays produced by the \co{} source to be representative of an urban
    environment.} 

    \label{fig:laboratory_detector_array}
\end{figure}

\shortsection{Laboratory Datasets} The two laboratory datasets were acquired
with a \si{4}-detector array setup (shown in
Figure~\ref{fig:laboratory_detector_array}) and a radioactive source at varying
distances between 0.5 and \SI{3}{\meter}. Both datasets contain \si{125}
samples where a \SI{3}{\micro\curie} \co{} radioactive source is rotated
\ang{90} at (roughly) \ang{15} increments around the detectors.  L-Dataset
\si{1} has no obstructions and L-Dataset \si{2} contains concrete obstructions
at fixed locations. A summary of the data is presented in
Table~\ref{table:dataset_details}, and a photograph of the detector setup and
accompanying block diagram are shown in
Figures~\ref{fig:block_diagram} \&~\ref{fig:laboratory_detector_array}.

While 125 samples per dataset may seem small, it is both significant in this
context and sufficient. A single sample often requires approximately 5 minutes
to collect (\ie{} nearly 11 hours of data collection for one dataset).
Moreover, throughout this entire process, we regularly performed energy
calibrations on the detector, and periodically acquired separate background
radiation spectra to make the readings as accurate as possible\footnotemark.
Finally, some learning models (\eg{} SVMs) are performant on small datasets.
Thus, these relatively small laboratory datasets represent a challenge in
localizing radioactive sources when data may be severely limited.

\footnotetext{Detectors in reality experience what is known as ``gain
shift''--the energy spectrum for  radiation slowly changes overtime from a
variety of environmental factors. Additionally, the background radiation can
vary with time and location. For radioactive source search scenarios, a single
calibration is often sufficient, however, data collection for scientific use,
such as this work, requires recalibrating the detector for gain shift and
background radiation regularly to obtain accurate measurements.}

\begin{table}
\centering
    \resizebox{\columnwidth}{!}{%
        \begin{tabular}{cccccccc} 

        \toprule

        Dataset & \begin{tabular}{@{}c@{}}Obstruction \\ Size
            (\SI{}{\metre})\end{tabular} & \begin{tabular}{@{}c@{}}Obstruction
        \\ Location\end{tabular}& Angle (\si{\degree}) & Distance
            (\SI{}{\metre}) & Num.  Trials\\

        \midrule

        S-Dataset 1 &  &  & \numrange{0}{360} & \numrange{1}{15} & 72,000 \\

        S-Dataset 2 & \(1\times2\times5\) & Fixed & \numrange{0}{360} &
        \numrange{1}{15} & 72,000\\

        S-Dataset 3 & \(0.5\times2\times5\) & Moving & \numrange{0}{90} &
        \numrange{1}{15} & 27,000 \\

        L-Dataset 1 & & & \numrange{0}{90} & \numrange{0.5}{3} & 125 \\

        L-Dataset 2 & \(0.5\times2\times5\) & Fixed & \numrange{0}{90} &
        \numrange{0.5}{3} & 125 \\

        \bottomrule

        \end{tabular}}

    \caption{Dataset statistics for the experiments.}

    \label{table:dataset_details}
\end{table}

\subsection{Experiment Overview}

\begin{figure*}[t]
    \centering
    \begin{subfigure}[t]{0.9\textwidth}
        \centering
        \includegraphics[width=\textwidth]{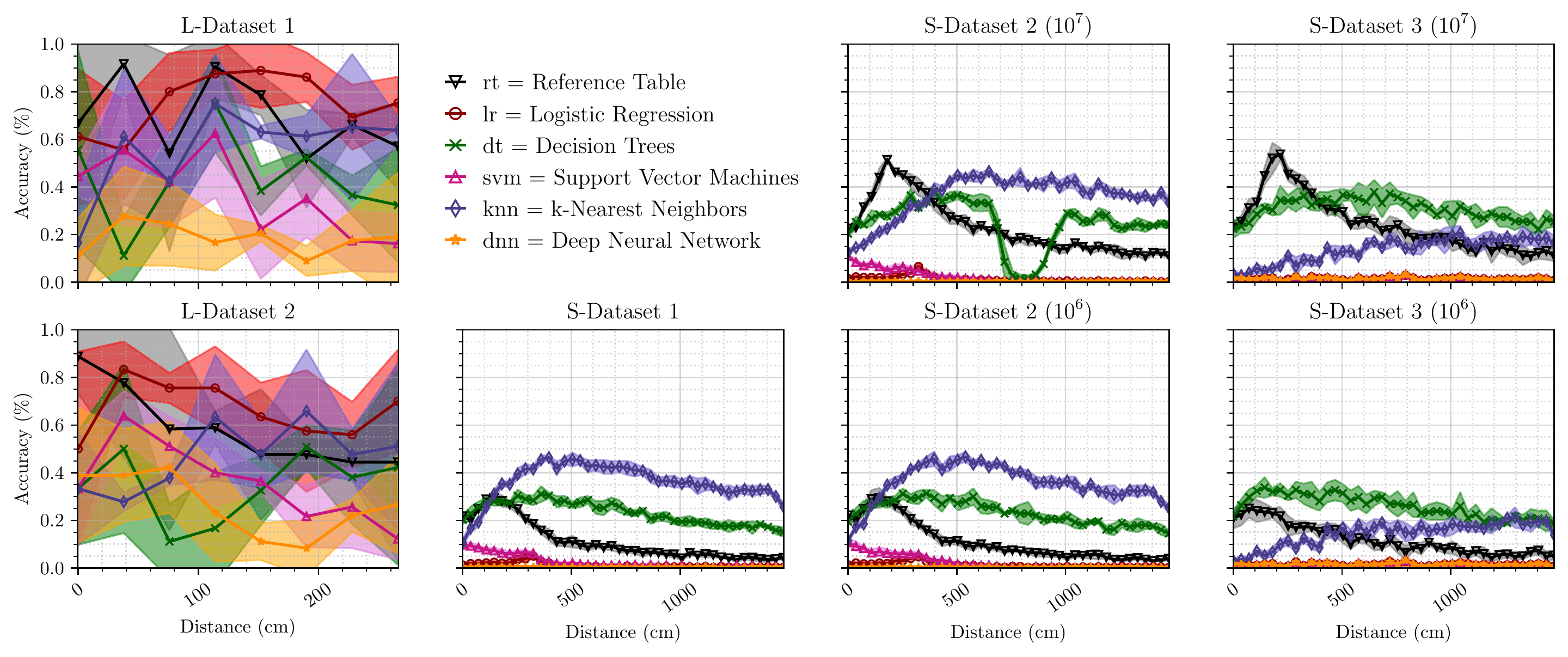}
        \caption{Predicting Angle}
    \end{subfigure}
    \begin{subfigure}[t]{0.9\textwidth}
        \centering
        \includegraphics[width=\textwidth]{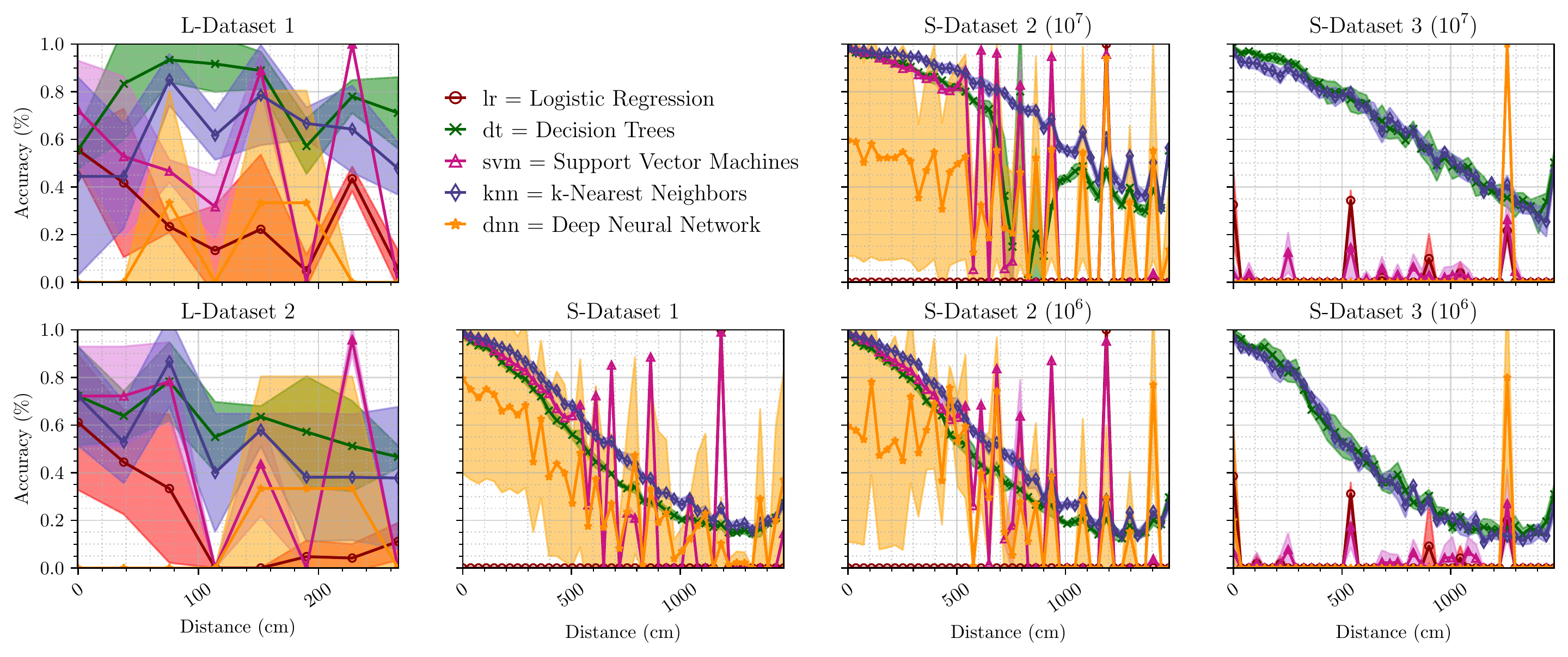}
        \caption{Predicting Distance}
    \end{subfigure}

    \caption{Localizing radioactive sources with naive machine learning -
    Applying machine learning to raw detector counts produced average results
    for predicting angle. Estimating distance was acceptable for some learning
    techniques.}

    \label{fig:raw_results}
\end{figure*}

This section details experiments exploring how the proposed models predict angle and distance as compared to the reference tables. In the following figures, the
accuracy is the number of samples where the exact angle (or
distance\footnotemark{}) was predicted correctly over the total number of
samples. We highlight these results as, in real scenarios, even \(\pm\ang{5}\)
angular tolerance may be unacceptable (particularly if the radioactive source
is estimated to be far away).

\footnotetext{The distances are binned (and approximated) via the
Freedman--Diaconis Estimator, which is an outlier-resilient, optimal binning
algorithm~\cite{freedman_histogram_1981}. Importantly, the estimator suggests
bin widths so that the difference between the empirical and theoretical
probability distributions are minimal. Reported accuracy is the number of
samples where the bin was predicted over the total number of samples. For the
simulations and laboratory experiments, 42 and 8 bins were created, \ie{}
\si{35} and \SI{3.75}{\centi\metre} per bin, respectively.}

\begin{figure*}[t]
    \centering
    \begin{subfigure}[t]{0.9\textwidth}
        \centering
        \includegraphics[width=\textwidth]{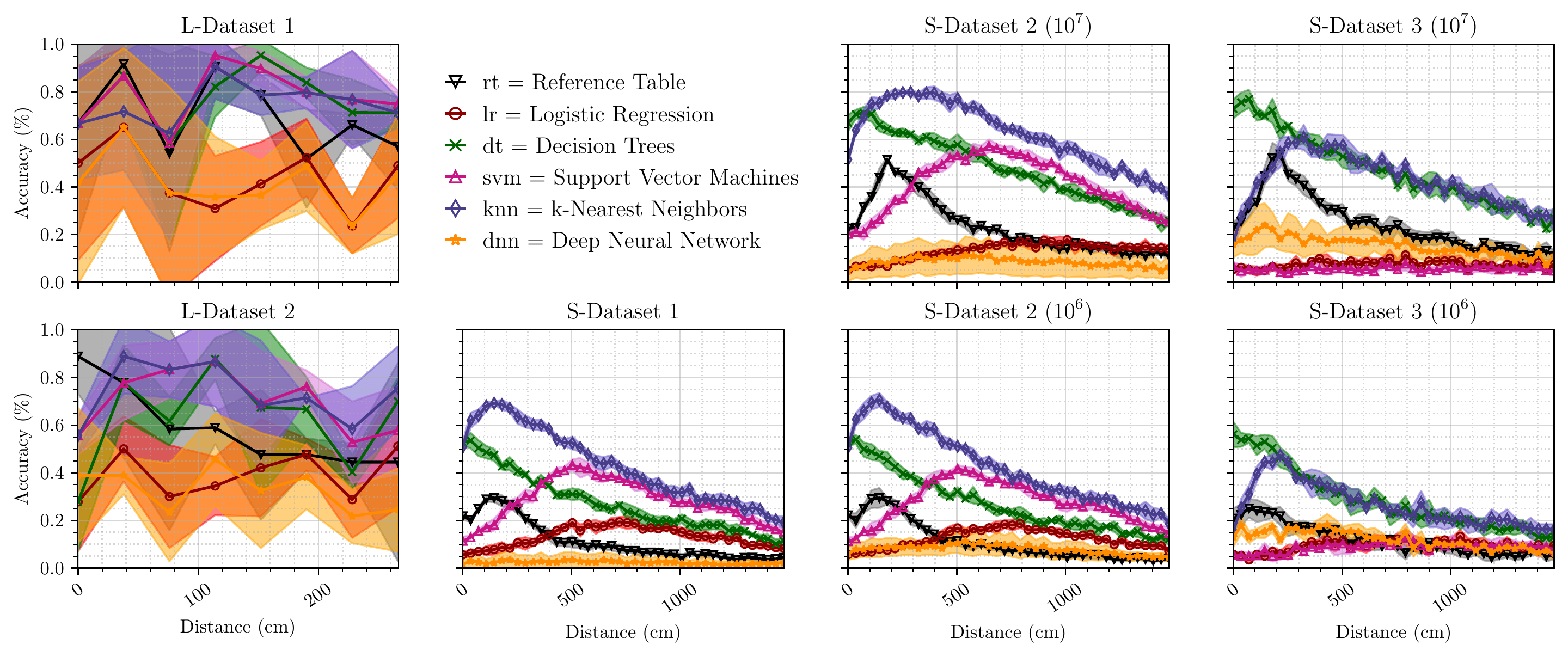}
        \caption{Predicting Angle}
    \end{subfigure}
    \begin{subfigure}[t]{0.9\textwidth}
        \centering
        \includegraphics[width=\textwidth]{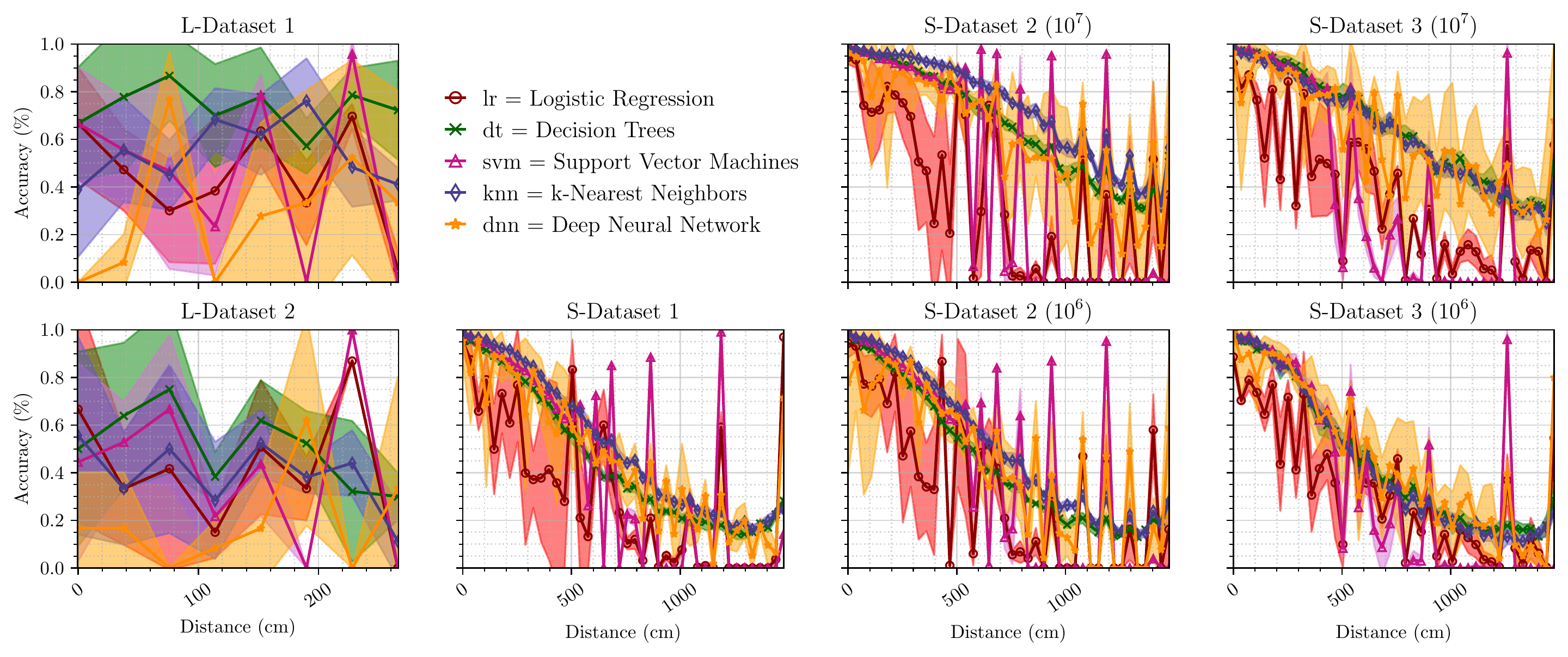}
        \caption{Predicting Distance}
    \end{subfigure}

    \caption{Localizing radioactive sources with cyber-security detection
    techniques - applying unit norm scaling and robust outlier standardization
    lead to significant improvements for most of the models.}

    \label{fig:unit_norm_and_robust_results}
\end{figure*}

\shortsection{Naive Machine Learning} The naive approach to this problem is to
directly use the raw counts collected by the detector as inputs to learning
models. As shown in Figure~\ref{fig:raw_results} (a), the results are average:
the reference table had average angular error of \ang{\rte}\(\rtesd\) while
the best model (k-Nearest Neighbor) had an average angular error of
\ang{\mlre}\(\mlresd\).The 
logistic regression model and deep neural network were unable to predict neither angle
nor distance correctly. For the distance at which the reference table was
calibrated (around \SI{200}{\centi\metre}; where the peaks are), the reference
table outperformed all of the models. As the distance increased, the reference
table quickly became inaccurate, unlike Decision Trees and k-Nearest Neighbors.
However, neither of these algorithms eclipsed the accuracy of the reference
table at any particular distance compared to the maximal accuracy of the
reference table (\ie{} the distance it was calibrated for). For the laboratory
experiments, the best models performed \textit{worse}: the reference table had
an average angular error of \ang{\lrte}\(\lrtesd\), while the best model
had an average angular error of \ang{\lmlre}\( \lmlresd\). Thus, for
predicting angle, the simple application of machine learning yields results
worse than reference tables at their calibrated distance, marginally better at
other distances for the simulated experiments, and explicitly worse for the
measured datasets.

For predicting distance (Figure~\ref{fig:raw_results} (b)), the naive approach
produced impressive results with some of the algorithms for scenarios in which the source strength is assumed to be known. The best model (k-Nearest Neighbors) could predict
distance within \SI{\mldre}{\percent}\(\mldresd\) of the distance to a
source. However, for the simulated data, most of the learning techniques were not able to
predict distance at all, while decision trees and k-Nearest Neighbor produced a
sigmoid-like curve for the \co{} \(10^6\) simulation.  Perhaps not
surprisingly, these models can estimate (nearly perfectly) the distance to
sources that are exceedingly close, yet struggle for sources that are
relatively far away. 

\shortsection{Applying Cyber-security Detection Techniques} As detailed in
Section~\ref{approach}, the radioactive environment is inherently burdened by
noise, similar to intrusion detection domains, and thus we suspected that a
cyber detection approach would readily apply in this domain.
Figure~\ref{fig:unit_norm_and_robust_results} (a) demonstrates the results.
Immediately, we can see significant improvements: many of the models now exceed
the reference table accuracy, even at the distance at which the
reference table was calibrated. After applying unit norm scaling, the average
angular error for the best model (k-Nearest Neighbors) is \ang{\mle}\(
\mlesd\) (down from \ang{\mlre}), an improvement from the reference table by
\SI{\error}{\percent} (down from \ang{\rte}). For the laboratory datasets, we
also see improvements: we reduce the angular error to \ang{\lmle}\(
\lmlesd\) (down from \ang{\lmlre}), an improvement from the reference table by
\SI{\lerror}{\percent}.

For predicting distance, we applied robust feature standardization. Like
network intrusion detection, this domain is inherently noisy and contains
outliers that may negatively influence standard feature scaling techniques.
Figure~\ref{fig:outliers} shows a sample distribution of detector counts with
the quartile ranges we scale from in the experiments and
Figure~\ref{fig:unit_norm_and_robust_results} (b) shows the results. A small improvement is made to the overall accuracy after applying robust feature
standardization (from \SI{\mldre}{\percent}\(\mldresd\) to
\SI{\mlde}{\percent}\(\mldesd\) for the best models) and substantial gains
for the other models (\eg{} logistic regression, support vector machines, and
deep neural networks) as shown in
Figure~\ref{fig:unit_norm_and_robust_results}. For the laboratory experiments,
robust feature standardization also yielded small improvements (from
\SI{\lmldre}{\percent}\(\lmldresd\) to \SI{\lmlde}{\percent}\(
\lmldesd\)). 

\vspace{3pt}
\noindent
We highlight key takeaways of this work:

\vspace{-0.75em}
\begin{itemize}

    \item Our approach can far surpass the capabilities of reference tables
        \textit{even for the distance at which the table was calibrated}.  This
        demonstrates that: \si{1}) calibrations do not lend themselves well to
        the complex nature of problems in real environments, and \si{2})
        model-based approaches, paired with cyber-security detection
        techniques, are effective tools for localizing radioactive sources.

    \item Our approach accurately estimates distance to a radioactive source.
        Prior to this work, techniques either required mobile detectors (either
        first responders on foot or vehicles) to triangulate radioactive
        sources manually; now, we are void of these limitations. We can
        localize a radioactive source simultaneously at the time it is
        detected. While initial trials benefited from an apriori knowledge of
        the source strength, similar standardization and additional
        cyber-physical security techniques are being explored to apply distance
        predictions to sources of unknown strength. 

    \item Perhaps not surprisingly, obstructions have a tangible impact on modeling radioactive behavior:
        most of the models observed an \({\sim}10\%\) decrease in accuracy in
        the most challenging datasets where the locations of obstructions
        varied. Radioactive source search scenarios in reality will likely
        observe similar challenges given that no environment is identical.

    \item An order of magnitude increase in particle counts (\(10^6\) to
        \(10^7\)) was especially helpful to increase model accuracy at the
        longest distances (\ie{} greater than \SI{8.4}{\metre}).  In other
        words, high activity sources (or longer acquisition times) can be
        localized accurately over long distances.

\end{itemize}

\section{Observations} \label{discussion}

\shortsection{Unit Norm Scaling} One of the most significant improvements we
observed for angle prediction was the application of unit norm scaling. There
are multiple reasons why this technique was so effective: much like detecting
spam in emails, the absolute frequency of words is hardly useful; instead, it
is often more interesting to see how frequent some words are used
\textit{relative to one another}. The intuition is straightforward: if the bulk
of an email contains words that are commonly associated with spam, then the
email is most likely spam as well; that is to say, long emails that contain
\SI{80}{\percent} ``spam words,'' for example, are fundamentally no different
(in terms of spam or not) than short emails with a similar relative amount of
spam words. We follow this same reasoning for localizing radioactive materials:
when receiving a sum total of \si{1000} counts or \si{100} counts, if a
particular detector receives the majority then, in both cases, the radioactive
source is most likely in front of this particular detector. This has the added
benefit of augmenting the training set--learning approaches no longer have to
disentangle that \si{1000} counts or \si{100} is relatively meaningless for
predicting angle, as those two situations are being treated identically. These
observations give insight into why this feature scaling technique was so
effective.

\begin{figure}[t]
    \includegraphics[width=\columnwidth]{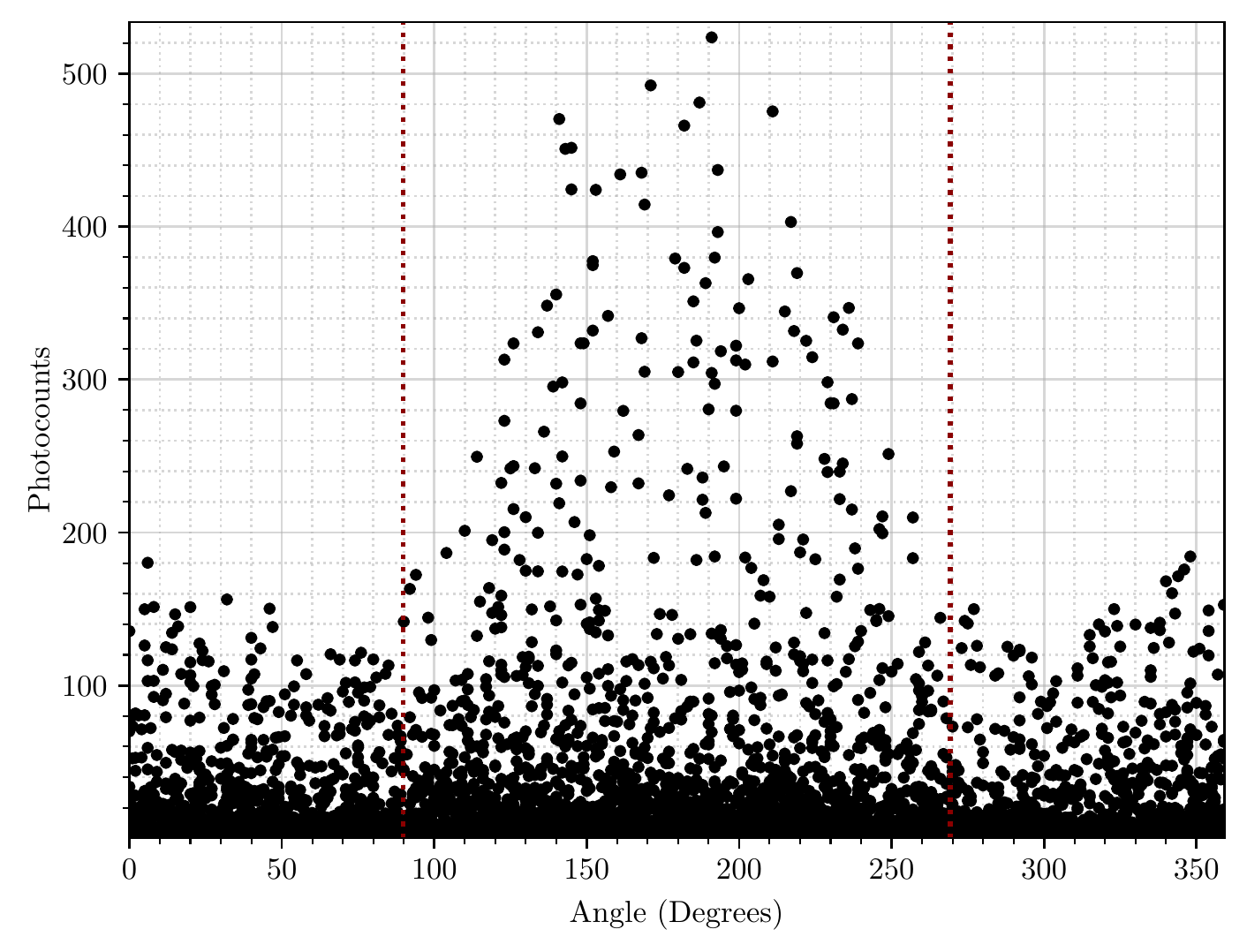}

    \caption{Demonstrating Outliers - This figure demonstrates a photocount
    distribution for one of the detectors. The dotted lines mark the quantile
    ranges of the data that we scale to. This gives intuition into why robust
    feature standardization is helpful: in this figure, the radioactive source
    is directly behind the detector (as shown by the peak at \ang{180}), yet
    there is a noticeable increase in photocount readings directly in front of
    the detector (\ie{} \ang{0} and \ang{359})--these increased counts are
    manifestations of noise in part from the physical phenomena.}

    \label{fig:outliers}
\end{figure}

\shortsection{Robust Outlier Standardization} We found that scaling the 
features in a manner robust to outliers was effective for predicting distance.
While using the raw counts was acceptable for sources that were close to the
detectors, we noticed that the accuracy of the models decreased quickly as the
distance linearly increased (\ie{} the Inverse-square law in practice). We
observed that robustly scaling features helped maintain the accuracy of the 
models over longer distances. Figure~\ref{fig:outliers} led us to our insight:
we aim to emphasize the signal from within the two dotted lines as the bulk of
the counts indicated that the source was directly behind the detector in this
example.  However, due, in part, to the physical phenomena described in
Section~\ref{radiation}, the detector observed a small increase in counts
directly in front of it (\ie{} at \ang{0}). Thus, we hypothesized that
mitigating the influence of these outliers would aid in predicting distance.
The results demonstrate that this insight was indeed helpful.

\shortsection{Estimating Distance} While these approaches are relatively accurate
at predicting distance, estimating distance is difficult, especially with a
stationary system. Today, there is a focus on using mobile systems to localize radioactive sources (e.g., Mobile Urban Radiation Search
(MURS)~\cite{curtis_simulation_2020}). By taking multiple samples
at different locations, mobile systems can exploit basic triangulation algorithms
to estimate the distance to a source (perhaps with more accuracy than our initial
approaches). A natural limitation of this approach is the cost, requirement
of a mobile environment, and dependency on multiple samples for triangulation.
However, the fact that the success of our approaches are agnostic to these 
requirements yields a unique opportunity; much like how predictions with 
the stationary system studied benefited from the application of cyber techniques,
the same is likely true for the techniques used in systems such as MURS. 
Broadly speaking, the 
overlap between radiation detection and detection in computing 
environments suggests that a large class of approaches are likely to benefit
from cyber-inspired approaches, as observed with the system used in this work.
\section{Conclusion} \label{conclusion}

In this paper, we investigated new analysis approaches for radioactive source localization. 
We explored how techniques from the cyber-security detection domains
can surpass traditional table-based approaches and extend the standard
definition of localization to include distance. We observed through both 
simulated and physical laboratory experiments that our techniques surpassed
the angular accuracy of table-based approaches, reducing the angular error
by \SI{\error}{\percent} and reliably predicting distance within
\SI{\mlde}{\percent}.

Moreover, we observed how naive applications of machine
learning either produced inaccurate predictions or 
failed to eclipse the accuracy of table-based approaches at distances
for which the table was calibrated. Yet, curated applications, such as unit-norm 
scaling and robust standardization, can produce results which surpass table-based
approaches across all evaluated distances. This work demonstrates that the complex
signals produced during source localization efforts can benefit from
machine learning approaches with curated application of signal amplification 
techniques. Future efforts will focus on more robust distance prediction techniques 
and greater collection of physical measurements in controlled laboratory settings.
Through our application of cyber-security principles, we introduced
a state-of-the-art approach in localizing radioactive sources in complex
physical scenarios.
\bibliographystyle{plain}
\bibliography{references}

\begin{thebibliography}{10}

\bibitem{noauthor_nuclear_nodate}
Nuclear {Development} : {Nuclear} {Energy} {Today}.
\newblock page 114.

\bibitem{altman_introduction_1992}
N.~S. Altman.
\newblock An {Introduction} to {Kernel} and {Nearest}-{Neighbor}
  {Nonparametric} {Regression}.
\newblock {\em The American Statistician}, 46(3):175--185, 1992.

\bibitem{boser_training_1992}
Bernhard~E. Boser, Isabelle~M. Guyon, and Vladimir~N. Vapnik.
\newblock A {Training} {Algorithm} for {Optimal} {Margin} {Classifiers}.
\newblock In {\em Proceedings of the {Fifth} {Annual} {Workshop} on
  {Computational} {Learning} {Theory}}, {COLT} ’92, pages 144--152, New York,
  NY, USA, 1992. Association for Computing Machinery.
\newblock event-place: Pittsburgh, Pennsylvania, USA.

\bibitem{casanovas_temperature_2012}
R.~Casanovas, J.J. Morant, and M.~Salvadó.
\newblock Temperature peak-shift correction methods for {NaI}({Tl}) and
  {LaBr3}({Ce}) gamma-ray spectrum stabilisation.
\newblock {\em Radiation Measurements}, 47(8):588--595, August 2012.

\bibitem{curtis_simulation_2020}
Joseph~C. Curtis, Reynold~J. Cooper, Tenzing~H. Joshi, Bogdan Cosofret, Thomas
  Schmit, John Wright, Jonathan Rameau, Daisei Konno, Daniel Brown, Forrest
  Otsuka, Eric Rappeport, Matthew Marshall, and Julia Speicher.
\newblock Simulation and validation of the {Mobile} {Urban} {Radiation}
  {Search} ({MURS}) gamma-ray detector response.
\newblock {\em Nuclear Instruments and Methods in Physics Research Section A:
  Accelerators, Spectrometers, Detectors and Associated Equipment}, 954:161128,
  February 2020.

\bibitem{durbin_development_2019}
Matthew Durbin, Ryan Sheatsley, Christopher Balbier, Tristan Grieve, Patrick
  McDaniel, and Azaree Lintereur.
\newblock Development of {Machine} {Learning} {Algorithms} for {Directional}
  {Gamma} {Ray} {Detection}.
\newblock In {\em Proceedings of the {Institute} of {Nuclear} {Materials}
  {Management}}, June 2019.

\bibitem{fine_feedforward_1999}
Terrence~L. Fine, S.~L. Lauritzen, M.~Jordan, J.~Lawless, and V.~Nair.
\newblock {\em Feedforward {Neural} {Network} {Methodology}}.
\newblock Springer-Verlag, Berlin, Heidelberg, 1st edition, 1999.

\bibitem{freedman_histogram_1981}
David Freedman and Persi Diaconis.
\newblock On the histogram as a density estimator:{L2} theory.
\newblock {\em Zeitschrift für Wahrscheinlichkeitstheorie und Verwandte
  Gebiete}, 57(4):453--476, December 1981.

\bibitem{goorley_initial_2013}
John~T. Goorley, Michael~R. James, Thomas~E. Booth, Jeffrey~S. Bull,
  Lawrence~J. Cox, Joe W.~Jr. Durkee, Jay~S. Elson, Michael~Lorne Fensin,
  Robert A.~III Forster, John~S. Hendricks, H.~Grady~III Hughes, Russell~C.
  Johns, Brian~C. Kiedrowski, Roger~L. Martz, Stepan~G. Mashnik, Gregg~W.
  McKinney, Denise~B. Pelowitz, Richard~E. Prael, Jeremy~Ed Sweezy, Laurie~S.
  Waters, Trevor Wilcox, and Anthony~J. Zukaitis.
\newblock Initial {MCNP6} {Release} {Overview} - {MCNP6} version 1.0.
\newblock Technical Report LA-UR-13-22934, 1086758, June 2013.

\bibitem{grypp_design_2014}
Matthew~D. Grypp, Craig~M. Marianno, John~W. Poston, and Gentry~C. Hearn.
\newblock Design of a spreader bar crane-mounted gamma-ray radiation detection
  system.
\newblock {\em Nuclear Instruments and Methods in Physics Research Section A:
  Accelerators, Spectrometers, Detectors and Associated Equipment}, 743:1 -- 4,
  2014.

\bibitem{hanna_directional_2015}
D.~Hanna, L.~Sagnières, P.J. Boyle, and A.M.L. MacLeod.
\newblock A directional gamma-ray detector based on scintillator plates.
\newblock {\em Nuclear Instruments and Methods in Physics Research Section A:
  Accelerators, Spectrometers, Detectors and Associated Equipment}, 797:13--18,
  October 2015.

\bibitem{humayed_cyber-physical_2017}
A.~Humayed, J.~Lin, F.~Li, and B.~Luo.
\newblock Cyber-{Physical} {Systems} {Security}—{A} {Survey}.
\newblock {\em IEEE Internet of Things Journal}, 4(6):1802--1831, December
  2017.

\bibitem{international_atomic_energy_agency_gamma_2006}
Industrial Applications {and} Chemistry~Section International Atomic
  Energy~Agency, Vienna~(Austria).
\newblock Gamma irradiators for radiation processing.
\newblock Technical Report 83-909690-6-8, IAEA, International Atomic Energy
  Agency (IAEA), 2006.
\newblock INIS-XA--862.

\bibitem{johns_physics_1983}
Harold~Elford Johns and John~Robert Cunningham.
\newblock {\em The physics of radiology}.
\newblock Charles C. Thomas, Springfield, Ill., U.S.A, 4th ed edition, 1983.

\bibitem{klann_current_2009}
Raymond~T. Klann, Jason Shergur, and Gary Mattesich.
\newblock Current {State} of {Commercial} {Radiation} {Detection} {Equipment}
  for {Homeland} {Security} {Applications}.
\newblock {\em Nuclear Technology}, 168(1):79--88, October 2009.

\bibitem{knoll_radiation_2000}
Glenn~F. Knoll.
\newblock {\em Radiation detection and measurement}.
\newblock Wiley, New York, 3rd ed edition, 2000.

\bibitem{kubat_machine_nodate}
Miroslav Kubat, Robert~C Holte, and Stan Matwin.
\newblock Machine {Learning} for the {Detection} of {Oil} {Spills} in
  {Satellite} {Radar} {Images}.
\newblock page~21.

\bibitem{malkoske_cobalt-60_nodate}
G~R Malkoske and J~Slack.
\newblock {COBALT}-60 {PRODUCTION} {IN} {CANDU} {POWER} {REACTORS}.
\newblock page~6.

\bibitem{mukhopadhyay_networked_nodate}
Sanjoy Mukhopadhyay, Richard Maurer, Ron Wolff, Ethan Smith, Paul Guss, and
  Stephen Mitchell.
\newblock Networked {Gamma} {Radiation} {Detection} {System} for {Tactical}
  {Deployment}.
\newblock page~9.

\bibitem{congress_of_the_united_states_congressional_budget_office_scanning_2016}
CONGRESS OF THE UNITED STATES CONGRESSIONAL~BUDGET OFFICE.
\newblock Scanning and {Imaging} {Shipping} {Containers} {Overseas}: {Costs}
  and {Alternatives}.
\newblock Technical report, CBO, June 2016.

\bibitem{ortiz_lessons_nodate}
P~Ortiz, M~Oresegun, and J~Wheatley.
\newblock Lessons from {Major} {Radiation} {Accidents}.
\newblock page~10.

\bibitem{pedregosa_scikit-learn_2011}
F.~Pedregosa, G.~Varoquaux, A.~Gramfort, V.~Michel, B.~Thirion, O.~Grisel,
  M.~Blondel, P.~Prettenhofer, R.~Weiss, V.~Dubourg, J.~Vanderplas, A.~Passos,
  D.~Cournapeau, M.~Brucher, M.~Perrot, and E.~Duchesnay.
\newblock Scikit-learn: {Machine} {Learning} in {Python}.
\newblock {\em Journal of Machine Learning Research}, 12:2825--2830, 2011.

\bibitem{protic_anomaly-based_2018}
Danijela Protić.
\newblock Anomaly-{Based} {Intrusion} {Detection}: {Feature} {Selection} and
  {Normalization} {Influence} to the {Machine} {Learning} {Models} {Accuracy}.
\newblock 1(3):7, 2018.

\bibitem{quinlan_induction_1986}
J.~R. Quinlan.
\newblock Induction of decision trees.
\newblock {\em Machine Learning}, 1(1):81--106, March 1986.

\bibitem{reeder_performance_nodate}
P~L Reeder and D~C Stromswold.
\newblock Performance of {Large} {NaI}({Tl}) {Gamma}-{Ray} {Detectors} {Over}
  {Temperature} -{50ºC} to +{60ºC}.
\newblock page~46.

\bibitem{remick_u.s._2005}
Alan~L. Remick, John~L. Crapo, and Charles~R. Woodruff.
\newblock U.{S}. {NATIONAL} {RESPONSE} {ASSETS} {FOR} {RADIOLOGICAL}
  {INCIDENTS}:.
\newblock {\em Health Physics}, 89(5):471--484, November 2005.

\bibitem{saha_physics_2001}
Gopal~B Saha.
\newblock {\em Physics and {Radiobiology} of {Nuclear} {Medicine}}.
\newblock 2001.

\bibitem{schrage_low-power_2013}
Chris Schrage, Nathan Schemm, Sina Balkir, Michael~W. Hoffman, and Mark Bauer.
\newblock A {Low}-{Power} {Directional} {Gamma}-{Ray} {Sensor} {System} for
  {Long}-{Term} {Radiation} {Monitoring}.
\newblock {\em IEEE Sensors Journal}, 13(7):2610--2618, July 2013.

\bibitem{sommer_outside_2010}
Robin Sommer and Vern Paxson.
\newblock Outside the {Closed} {World}: {On} {Using} {Machine} {Learning} for
  {Network} {Intrusion} {Detection}.
\newblock In {\em 2010 {IEEE} {Symposium} on {Security} and {Privacy}}, pages
  305--316, Oakland, CA, USA, 2010. IEEE.

\bibitem{kotagiri_robust_2007}
Jungsuk Song, Hiroki Takakura, Yasuo Okabe, and Yongjin Kwon.
\newblock A {Robust} {Feature} {Normalization} {Scheme} and an {Optimized}
  {Clustering} {Method} for {Anomaly}-{Based} {Intrusion} {Detection} {System}.
\newblock In Ramamohanarao Kotagiri, P.~Radha Krishna, Mukesh Mohania, and
  Ekawit Nantajeewarawat, editors, {\em Advances in {Databases}: {Concepts},
  {Systems} and {Applications}}, volume 4443, pages 140--151. Springer Berlin
  Heidelberg, Berlin, Heidelberg, 2007.

\bibitem{tahir_efficacy_2019}
Muhammad Tahir, Mingchu Li, Naeem Ayoub, and Muhammad Aamir.
\newblock Efficacy {Improvement} of {Anomaly} {Detection} by {Using}
  {Intelligence} {Sharing} {Scheme}.
\newblock {\em Applied Sciences}, 9(3):364, January 2019.

\bibitem{tsai_intrusion_2009}
Chih-Fong Tsai, Yu-Feng Hsu, Chia-Ying Lin, and Wei-Yang Lin.
\newblock Intrusion detection by machine learning: {A} review.
\newblock {\em Expert Systems with Applications}, 36(10):11994--12000, December
  2009.

\bibitem{u.s.nrc_nrc:_nodate}
U.S.NRC.
\newblock {\em {NRC}: {Radiation} {Basics}}.

\bibitem{walker_epa_nodate}
Stuart Walker.
\newblock {EPA} {Facts} about {Cobalt}-60.
\newblock Technical report, U.S. EPA.

\bibitem{wright_logistic_1995}
Raymond~E. Wright.
\newblock Logistic regression.
\newblock In {\em Reading and understanding multivariate statistics.}, pages
  217--244. American Psychological Association, Washington, DC, US, 1995.

\bibitem{ziock_lost_2002}
K.~P. Ziock.
\newblock The {Lost} {Source}, {Varying} {Backgrounds} and {Why} {Bigger} {May}
  {Not} {Be} {Better}.
\newblock In {\em {AIP} {Conference} {Proceedings}}, volume 632, pages 60--70,
  Washington, DC (USA), 2002. AIP.
\newblock ISSN: 0094243X.

\end{thebibliography}
\clearpage \appendix \section{Appendix} \label{appendix}
\begin{table*}[!]
    \centering
    \begin{tabular}{l p{6cm}}

        \toprule
        Logistic Regression & Solver: L-BFGS \newline
        Epochs: 100 \newline
        1.0 \(L_2\) 
        Regularization  \\
        \vspace{0.1mm} Decision Trees & Criterion: Gini \newline
        Max Depth: 10\\
        \vspace{0.1mm} Support Vector Machines & Kernel: RBF \newline 
        Gamma: 1/(\# features * variance) \newline 
        1.0 \(L_2\) Regularization \\
        \vspace{0.1mm} k-Nearest Neighbors & Neighbors: 5 \newline
        Algorithm: Ball Tree (Leaf size of 30) \newline 
        Distance: Minkowski \\
        \vspace{0.1mm} Deep Neural Network & Layers: 3 with 15 neurons (fully-connected) \newline Activation: Relu \newline
        Epochs: 300  \newline
        Optimizer: Adam \newline
        \(10^{-4}\) \(L_2\) Regularization  \\
        \bottomrule
    \end{tabular}
    \caption{Model Hyperparameters}

    \label{tab:ml}
\end{table*}

We provide a brief summary of the evaluated learning techniques and show our 
hyperparameters used in the evaluation. 

\shortsection{Logistic Regression} Logistic regression (LR), akin to linear
regression, computes a weighted sum of input features with an additional bias
term, and applies the logistic function to this
sum~\cite{wright_logistic_1995}. We can define a logistic regression model as
follows:

\begin{equation} \label{eq:lr}
    \ell = (1+e^{(-\theta^\top\cdot{x})})^{-1}
\end{equation}

\noindent where $\theta$ represents the weights of the model and $x$ represents
our input vector. While more sophisticated techniques have emerged, \ie deep
learning, each have their own limitations. We include logistic regression models in this 
work to investigate if simpler models suffice to perform localization tasks
accurately and quickly

\shortsection{Support-vector Machines} Prior to the inception of deep learning,
support-vector machines (SVMs) dominated machine learning benchmarks across
many domains~\cite{boser_training_1992}. For binary classification, 
SVMs seek to find two hyper-planes that satisfy:

\begin{equation} \label{eq:svm}
    \theta^\top\cdot{x}+b \lesseqgtr \{-1, 1\}
\end{equation}

\noindent where \(\theta\) represents the weights of the model, \(x\) 
represents our input vector, and \(\{-1, 1\}\) encode the two classes.
The algorithm then seeks to maximize the difference between the two hyperplanes,
while satisifying the above constraints. SVMs are attractive as they can form
non-linear decision boundaries, which may be necessary given the noisiness of
this domain.

\shortsection{k-Nearest Neighbors} The k-nearest neighbors algorithm (kNN) is a
non-parametric approach, used ubiquitously in academia and industry. As the
observed count distributions in our domain can change rapidly in a variety of
unique environments, kNN is particularly useful as it does not make any
assumptions about the underlying data. We can define the kNN algorithm as
follows (using the Euclidean metric for distance):

\begin{equation} \label{eq:knn}
    \hat{y} = \argmin_{y:(x, y)\sim \mathcal{D}} \sqrt{(\hat{x}-x)^{2}}
\end{equation}

\noindent where $\hat{y}$ is predicted class, $(x, y)\sim \mathcal{D}$
represents the input-class pairs for the observed distribution, and $\hat{x}$
represents a new observation at the time of inference. As a non-parametric
model, we expected kNN to exhibit adequate accuracy, amortized across a variety
of unique scenarios.

\shortsection{Decision Trees} Decision Trees (DT) are flexible machine learning
algorithms that are commonly used today~\cite{quinlan_induction_1986}. They
require minimal data preparation, have low performance overheads, and offer
intuitive explanations of the formed decision boundaries. Decision trees often 
follow a binary if-then-else structure whose rules are built by minimizing:

\begin{equation} \label{eq:dt}
    T(i, t) = \frac{|x|:x_i \leq t}{|\mathcal{D}_T|}\cdot{G(i, \leq t)} + 
    \frac{|x|:x_i > t}{|\mathcal{D}_T|}\cdot{G(i, > t)}
\end{equation}

\noindent where $i$ a feature, $t$ is the threshold for $i$, $x$ are input
vectors from distribution $\mathcal{D}_T$ partitioned at node $T$, and $G(i,
t)$ (\ie{} the Gini impurity score) is:

\begin{equation*} \label{eq:gini}
    G(i, \lesseqgtr t) = 1-\sum_{y} \left(\frac{|x|:x_i \lesseqgtr t, 
    (x, y)\sim \mathcal{D}_T}{|x|:x_i \lesseqgtr t}\right)^2
\end{equation*}

\noindent where $(x, y)\sim \mathcal{D}_T$ represents the input-class pairs for
the observed distribution. Decision trees are appealing in this domain not only
as another non-parametric technique, but also because of their white-box
design: the learned decisions are easy to interpret, which can be useful in
understanding subtle changes in the decision process as a function of the
environment.

\shortsection{Deep Neural Networks} 
Deep neural networks represent a
state-of-the-art class of learning techniques that have demonstrated success in
the most challenging machine learning benchmarks~\cite{fine_feedforward_1999}.
Definitions vary, but generally speaking, deep neural networks often refer to
any class of artificial neural networks with multiple layers between the input
and output layers. We can formalize a deep neural network (with ReLU as the activation
function) as: 

\begin{equation} \label{eq:mlp}
    P(x) = S(P^\ell(\max(0, \theta_\ell^\top\cdot{x}+b)))
\end{equation}

\noindent where $x$ is an input vector, $P^\ell$ is the $\ell^{th}$ iterate of
$P$ (\ie{} function composition) where $\ell$ is the number of layers in the
network, $\theta_\ell$ are the weights for the $\ell^{th}$ layer, $b$ is a
vector of biases, and $S$ is defined as the softmax layer. Given the
computational complexity required of artificial neural networks with many
hidden layers, we interested if deep neural networks were expressive enough to 
be effective in our most complex scenarios.

\end{document}